 \documentclass[usletterpaper, 10pt, conference]{IEEEtran}
\IEEEoverridecommandlockouts

\usepackage{enumitem}

\usepackage{enumitem}
\usepackage{amsmath}
\usepackage{amssymb}
\usepackage{cite} 
\usepackage{booktabs}    
\usepackage{multirow}    
\usepackage{graphicx}    
\usepackage{siunitx}
\usepackage{array}
\usepackage{longtable}   
\usepackage{array}       
\usepackage{booktabs}   
\usepackage{float}      
\usepackage{placeins}
\usepackage{standalone}
\usepackage{subfiles}
\usepackage{rotating}   
\usepackage{pdflscape}  
\usepackage{pgfplotstable}
\usepackage{pgfplots}
\usepackage{amsmath}
\usepackage{svg}
\usepackage[top=1 in, bottom=0.75in, left=0.75in, right=0.75in]{geometry}
\pgfplotsset{compat=1.18}
\begin{document}

\title{V2F: Vision-Informed Grasp Force Prediction for Damage-Aware Robotic Handling of Date Fruits}

\author{Shahd Shami, Obadah Wali, Eric Feron, and Shinkyu Park
\thanks{The work was supported by funding from King Abdullah University
of Science and Technology (KAUST).}
\thanks{The authors are with the Computer, Electrical and Mathematical Sciences and Engineering, King Abdullah University of Science and Technology (KAUST), Thuwal 23955, Saudi Arabia. {\tt \{shahd.shami, obadah.wali, eric.feron, shinkyu.park\}@kaust.edu.sa}}
}

\maketitle

\IEEEpeerreviewmaketitle
\begin{abstract}
This paper presents a vision-informed grasp force prediction framework for robotic handling of date fruits. Addressing the dual challenge of high detachment forces and low bruise thresholds, we first conduct mechanical characterization on date samples to define a safe grasping envelope and quantify the relationship between fruit geometry and bioyield stress. In this work, we develop a \textit{Vision-to-Force (V2F)} pipeline that combines computer vision-based segmentation, active-contour refinement, and geometric feature extraction with a physics-informed residual neural network that augments a Hertz contact equation. 
The resulting model maps non-contact visual descriptors and cultivar
metadata to predict a safe grasp force with mean validation performance of $R^2 \approx 0.7$ across unseen cultivar groups, which is a 
good result given the inherent mechanical variability of 
biological tissue. Experimental validation using a gripper and load cell indicates that the predicted forces enable stable manipulation of different types of date fruits, with residual deformations below 1 mm and no observable damage. These results show that pre-emptive, vision-driven force estimation
, enabling safer robotic handling of fragile fruits.

\end{abstract}


\section{Introduction}

The demand for robotic solutions in agricultural handling and harvesting has surged as the industry seeks to overcome labor shortages and improve post-harvest quality \cite{BECHAR201694,bharad2024agricultural} and reduce labor risks. While significant progress has been made in automating the harvest of rigid fruits\cite{ESPINOZA2026101918,zhou2021tactile}, the manipulation of delicate organic products in unstructured environments remains a significant challenge. Among these, the date fruit (\textit{Phoenix dactylifera}) is a crop of high economic and cultural importance, particularly in the Middle East and Mediterranean regions. However, the high variability in fruit texture and fragility requires robust perception and force-aware manipulation, making date harvesting a challenging and relatively underexplored problem in agricultural robotics.
 
Current robotic harvesting systems predominantly target rigid fruits using uniform clamping forces or simplified vacuum suction mechanisms. While recent advancements have introduced specialized end-effector designs for soft-fruit manipulation \cite{elfferich2022soft,visentin2023soft,HU2022107177}, these solutions often depends on tactile exploration which can cause micro-bruising before it reacts to stop. Dates present a complex mechanical paradox: they require a relatively high gripping force to overcome the detachment resistance of the bunch, yet possess a low threshold for mechanical deformation, making them highly susceptible to bruising. Furthermore, existing research lacks a predictive framework that can pre-emptively determine the optimal grasping force based on individual fruit characteristics. This absence of a predictive mapping between visual fruit characteristics and safe grasping forces results in a lack of adaptive grasping strategies, often leading to fruit slippage or irreversible tissue damage during robotic handling. 

To train and evaluate the proposed framework, we constructed a laboratory-scale dataset comprising
date fruits samples spanning multiple cultivars and hydration states. Controlled mechanical characterization experiments established the relationship between fruit geometry and safe grasping limits, providing the supervisory labels for the proposed learning framework.

\begin{figure}[!h]
    \centering
    \includegraphics[width=\linewidth]{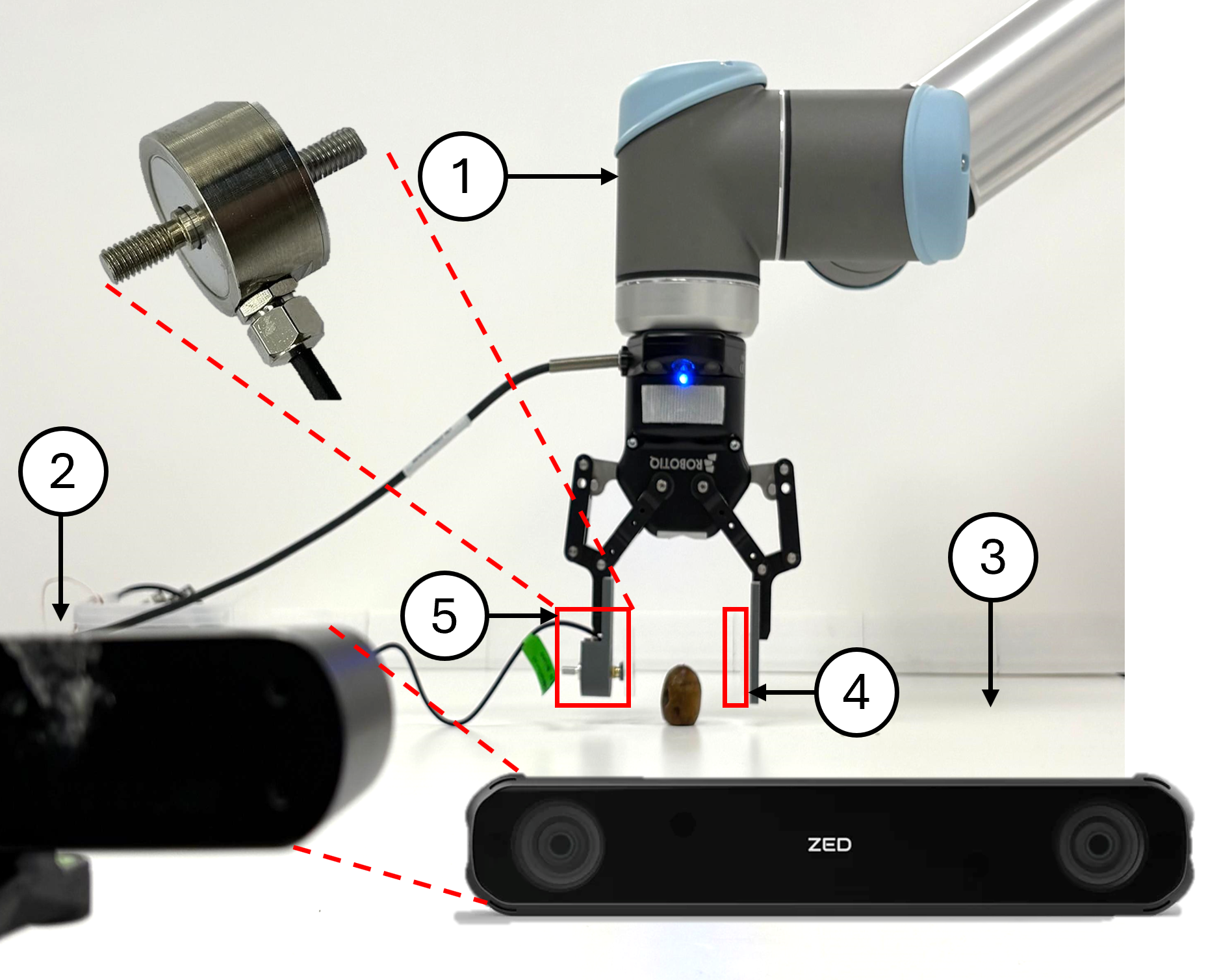}
    \caption{Experimental testbed for V2F validation: (1) UR10e robotic manipulator, (2) ZED stereo camera for visual feature extraction, (3) standardized testing surface, (4) custom acrylic gripper pads, and (5) high-precision load cell for real-time force feedback.}
    \label{fig:ur10}
\end{figure} 

Building upon these biomechanical insights, we propose Vision-to-Force (V2F), a framework that predicts safe grasping forces directly from pre-contact visual observations. V2F framework leverages computer vision and machine learning to estimate optimal grasping parameters. 
By mapping pre-contact visual descriptors to the required normal grasp force, the framework enables the robot to adapt its grip to the mechanical properties of each individual fruit while maintaining damage-free manipulation.


The primary contributions of this work are as follows:
\begin{itemize}
\item A Vision-to-Force (V2F) framework for predicting safe grasp forces directly from pre-contact visual observations.
\item A physics-guided residual learning model that combines Hertzian contact mechanics with data-driven learning to improve force prediction.
\item Experimental validation on a robotic gripper equipped with a force sensor (see Fig.~\ref{fig:ur10}), demonstrating the feasibility of using ML-driven force identification for the damage-free handling of delicate fruits.
\end{itemize} 
The remainder of this paper is organized as follows: Section~\ref{sec:related} reviews related work in agricultural manipulation. Section~\ref{sec:mechanical_modeling} details the mechanical characterization of the date fruit dataset. The proposed V2F ML framework is described in Section~\ref{sec:V2F}. Section~\ref{sec:Exp_validation} presents the robotic hardware integration and experimental results, and Section~\ref{sec:conclusion} concludes the paper with future research directions.



\section{Related Work}
\label{sec:related}

\subsection{Pre-Harvest Fruit Classification and Maturity Assessment}

Vision-based classification of date fruits prior to harvesting has seen 
significant progress through deep learning. Altaheri~et~al.~\cite{dcrhne} 
trained a CNN 
to classify cultivar and 
ripeness under natural lighting conditions.
Faisal~et~al.~\cite{IHDS} extended this direction with the Intelligent 
Harvesting Decision System (IHDS)
to classify maturity stages.
While these results are 
compelling, both studies operate exclusively in the visual domain and 
report no mechanism for translating maturity predictions into mechanically 
safe grasping decisions.
Therefore, visual classification alone cannot determine the grasp force required for safe manipulation.

\subsection{Grasp State Classification and Safe Force Inference}

While visual classification identifies harvest-ready fruits, safe manipulation additionally requires estimating the forces needed for stable grasping. Reliable grasping in cluttered, occluded environments requires sensors 
that can detect and respond to unstable contact events without relying 
on line-of-sight perception. Walt and Krishnan~\cite{Grasp_State} 
addressed this directly in the agricultural setting, mounting low-cost 
IMU and IR reflectance sensors on a two-finger gripper to classify four 
grasp states: no slip, slip, failed grasp, and successful pick, using 
a Random Forest classifier operating on FFT-transformed sensor windows. 
Their system achieved perfect recall on failed grasps and was validated 
on a live cherry tomato plant, demonstrating that computationally light 
sensing pipelines can function reliably under the occlusion and stem 
attachment forces unique to agricultural manipulation. A key finding was 
that sensor fusion, rather than any single modality, provided the 
robustness needed for real-world plant conditions.

Building on this, recent work has moved beyond state classification 
toward predicting the safe operating force for deformable objects. 
Liu and Sun~\cite{KAN_ViT} proposed KAN-ViT, an end-to-end transformer 
architecture that fuses GelSight tactile images with external RGB camera 
data to jointly classify grasp states and infer safe grasping force for 
soft objects such as bread. Their framework introduces optical flow 
extracted from tactile image sequences to capture subtle contact dynamics 
that static images miss, and replaces the standard MLP decoder with a 
KAN network to improve time efficiency. Force calibration is achieved 
through a 3D gel indentation mapping correlating visual sensor deformation 
with normal contact force. Xu~et~al.~\cite{AM_LSTM} further showed that 
an attention-augmented LSTM correctly up-weights informative contact 
frames from piezoresistive tactile sequences, and that gripper closing 
speed significantly affects signal quality and recognition accuracy,
rising above 90\% at high speed but falling below 71\% at the slowest 
tested speed, confirming that grasp dynamics, not just final contact 
geometry, shape the measurable signal.

Although these studies advance grasp-state monitoring and reactive force 
control, they operate exclusively during or after contact, 
classifying an ongoing grasp or adapting force in response to observed 
slip. None estimates the required force prior to contact. 
Furthermore, their dependence on high-resolution vision-based tactile 
sensors and online exploratory actions introduces infrastructure that 
may be impractical in field harvesting robots.

\subsection{Vision-Tactile Fusion and Deformable Object Grasping}

Beyond agricultural applications, multimodal vision-tactile methods have also been investigated for deformable object grasping.
Wang~et~al.~\cite{Gelsight_ICRA_2021} introduced a compact GelSight-based 
tactile sensor capable of reconstructing high-resolution 3D contact 
geometry, demonstrating that accurate contact modeling improves grasp 
interpretation for deformable objects. Kwiatkowski~et~al.~\cite{IROS_2017_Grasp_Stability} 
proposed a multimodal framework fusing proprioceptive and tactile signals 
to assess grasp stability, showing that sensor fusion significantly 
enhances robustness over single-modality systems. Slip detection has 
been widely investigated; Dong and Adelson~\cite{ICRA_2018_Slip} 
demonstrated that combining tactile and visual information improves 
slip prediction reliability under dynamic contact conditions. More 
recently, transformer-based architectures have been proposed for 
deformable object grasping, leveraging cross-modal attention to 
generalize across object categories and grasp 
configurations~\cite{IROS_2023_Deformable_Transformer}.

Across all of these works, force inference is derived from rich contact 
sensing during or after grasp interaction. Our framework 
shifts prediction earlier in the manipulation pipeline: rather than 
relying on high-resolution tactile imaging or exploratory probing, we 
estimate yield-level grasp stress prior to contact from compact 
geometric descriptors and fruit type, grounding predictions in analytical 
Hertzian contact mechanics and refining them through physics-guided 
residual learning. This positions our approach as a planning-time 
complement to the execution-stage feedback systems reviewed above.


\section{Mechanical Characterization and Force Modeling}

\label{sec:mechanical_modeling}

In this section, we establish the biomechanical foundation required to define the ``safe grasping envelope" for date fruits. This physical grounding ensures that our predictive model provides force set-points that maintain the fruit within its elastic deformation regime, avoiding irreversible tissue damage.

\subsection{Biomechanical Properties of \textit{Date Fruit}}
\begin{figure}[htbp]
    \centering
     \includegraphics[width=\linewidth]{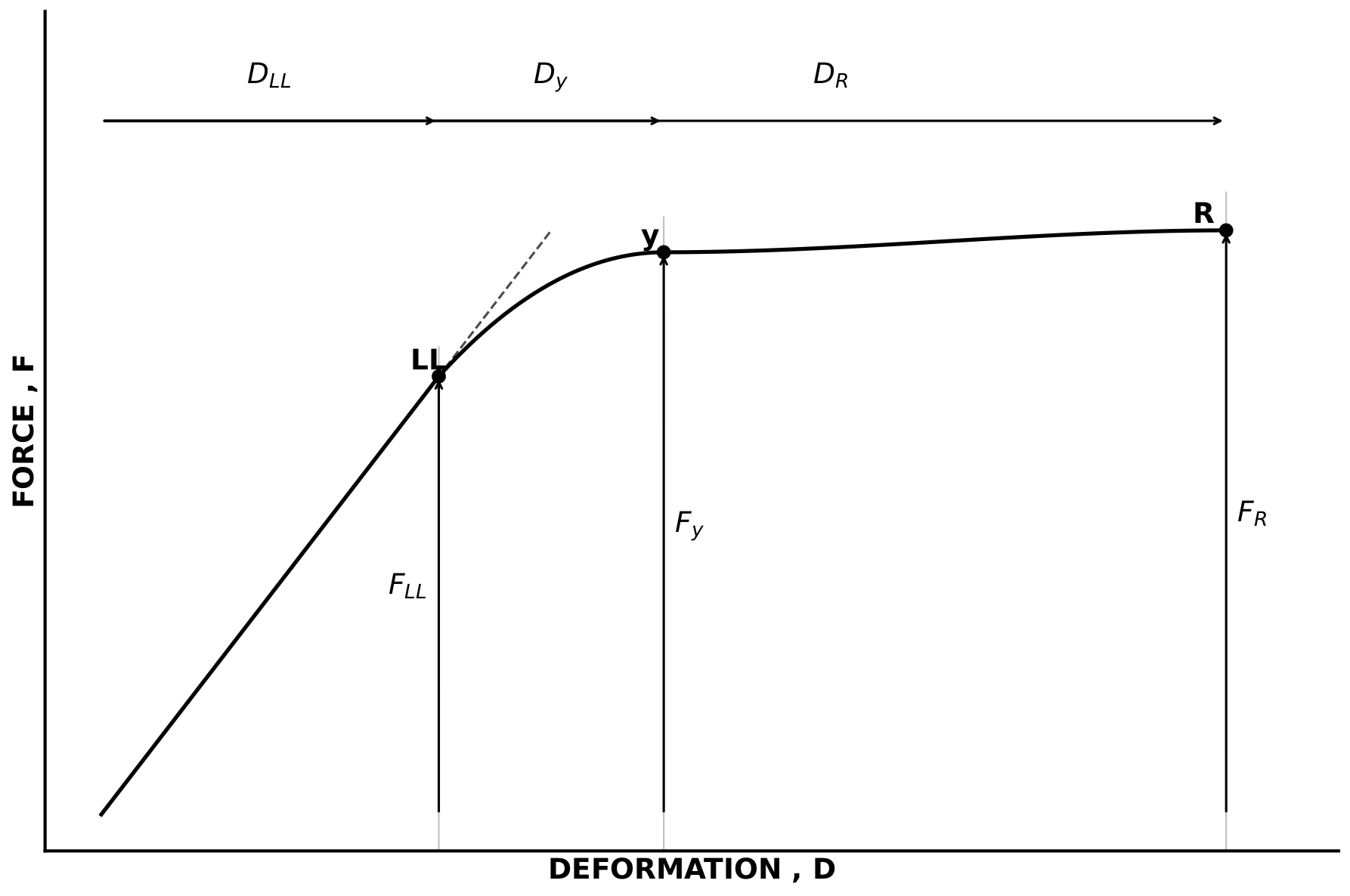}
    \caption{A typical force-deformation curve for an agricultural product, showing the proportional limit LL, bioyield point y, and rupture point R. Corresponding forces ($\text{F}$) and deformations ($\text{D}$) are labeled at each point $\in \{\text{LL}, \text{y}, \text{R}\}$~\cite{mohsenin2020physical}.}
    \label{fig:fvsd}
    \end{figure}

Agricultural products like dates exhibit complex viscoelastic behavior, where the stress-strain relationship is influenced by cellular structure, moisture content, and maturity \cite{mohsenin2020physical}. As illustrated in Fig. \ref{fig:fvsd}, the force-deformation curve of the date fruit is characterized by three critical transition points:
\begin{enumerate}
    \item Linear Limit (LL): The regime of proportional elasticity where the slope remains constant.
    \item Bioyield Point ($\mathrm y$): The onset of microstructural cell rupture. This represents the functional limit for damage-free robotic handling.
    \item Rupture Point ($\mathrm R$): Macrostructural failure, such as skin puncture or cracking.
\end{enumerate}

The mechanical response of the fruit is characterized using engineering stress and strain. The applied compressive stress is $ \sigma = F/A_0$, where $F$ is the applied compression force and $A_0$ is the initial contact area. The corresponding engineering strains along the loading ($y$) and transverse ($z$) directions are given by $\epsilon=\delta_L/L_0, $
where $L_0$ denotes the initial dimension, and $\delta_L$ represents the corresponding dimensional changes during compression.
The mechanical resistance of the fruit is governed by its Young’s modulus ($E = \sigma / \epsilon_y$) and Poisson's ratio ($\nu = -\epsilon_z / \epsilon_y$). These parameters determine the material’s shear modulus ($G$), defined as
\begin{equation}
    G = \frac{E}{2(1+\nu)}.
\end{equation}
Maintaining the applied robotic stress below the bioyield threshold $\sigma_{y}$ ensures that deformation remains elastic, preserving the fruit's commercial value.
      
Our objective is to determine a force $F_{\text{tot}}$ such that the maximum contact stress $\sigma_{\max}$ satisfies the safety condition:
\begin{equation}
    \sigma_{\max} < \sigma_{\text{bioyield}}.
\end{equation}

\subsection{Bridging Physics to ML: Feature Mapping}
The core hypothesis of our V2F framework is that a fruit's morphology, specifically its projected surface area, serves as a reliable proxy for its mechanical constraints. 
      
By identifying the fruit's projected area via computer vision, we can estimate its effective contact patch and geometry. This transitions the problem from a reactive tactile sensing task to a \textbf{predictive vision task}. The ML model, described in the following section, learns to map these visual descriptors to the optimal $F_{\text{grasp}}$.

\section{Vision-To-Force Framework}
\label{sec:V2F}

\begin{figure*}[!t]
    \centering
        \includegraphics[width=\linewidth]{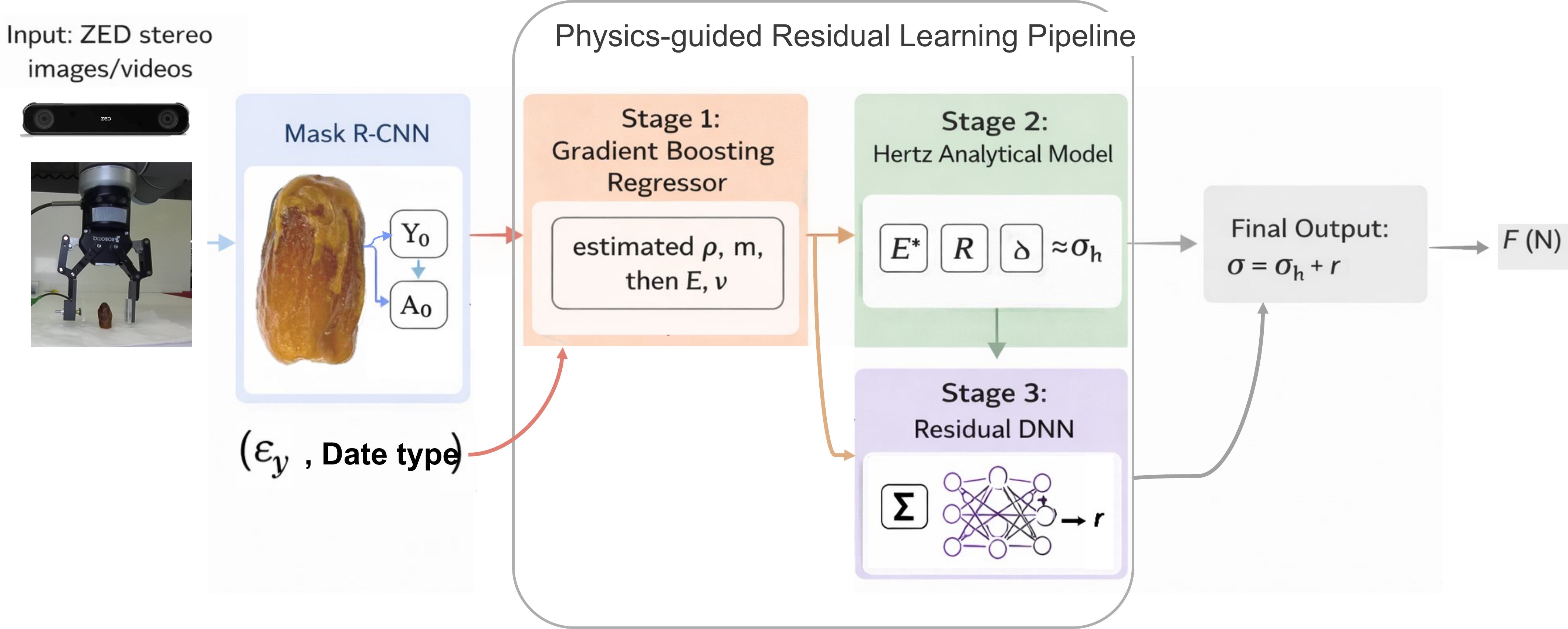}
\caption{ Architecture of the physics-guided residual learning framework. The pipeline integrates analytical Hertz's contact theory (Stage 2) with a deep residual network (Stage 3) to predict grasping stress by mapping visual geometric features to biomechanical material properties.}
\label{fig:framework_structure}
    \label{fig:SD}
\end{figure*}

The proposed V2F pipeline, illustrated in Fig. \ref{fig:framework_structure}, employs a three-stage physics-guided residual learning architecture to map visual descriptors to mechanical grasping constraints. The process initiates with stereo image acquisition to extract the initial fruit dimentions.
In Stage 1, a Gradient Boosting Regressor estimates the fundamental material properties.
Stage 2 utilizes these parameters within an analytical Hertz's contact model to compute the theoretical stress ($\sigma_h$)
Finally, in Stage 3, a residual deep neural network identifies the discrepancy ($r$) between the analytical model and the measured stress ($\sigma_{\mathrm{meas}}$). 

\subsection{Data and Inputs}
The training dataset was obtained under controlled laboratory conditions using date fruits at the Tamar ripeness stage, which represents the fully ripe stage commonly associated with marketable and consumable dates~\cite{Ahmed2014,Haider2018}. The samples included three Saudi cultivars, Ajwa, Barhi, and Sagai, prepared under normal, moistened, and dried conditions to capture practical variability in fruit texture and moisture state during handling. Mechanical labels were generated from compression tests, where force and deformation were recorded synchronously to identify safe grasping limits before the onset of permanent tissue damage. The dataset size was determined using Simple Random Sampling (SRS) methodology~\cite{lohr2021sampling},
which yielded approximately 500 samples.
\begin{figure}[htbp]
    \centering
    \includegraphics[
    height= 4 cm
    ]{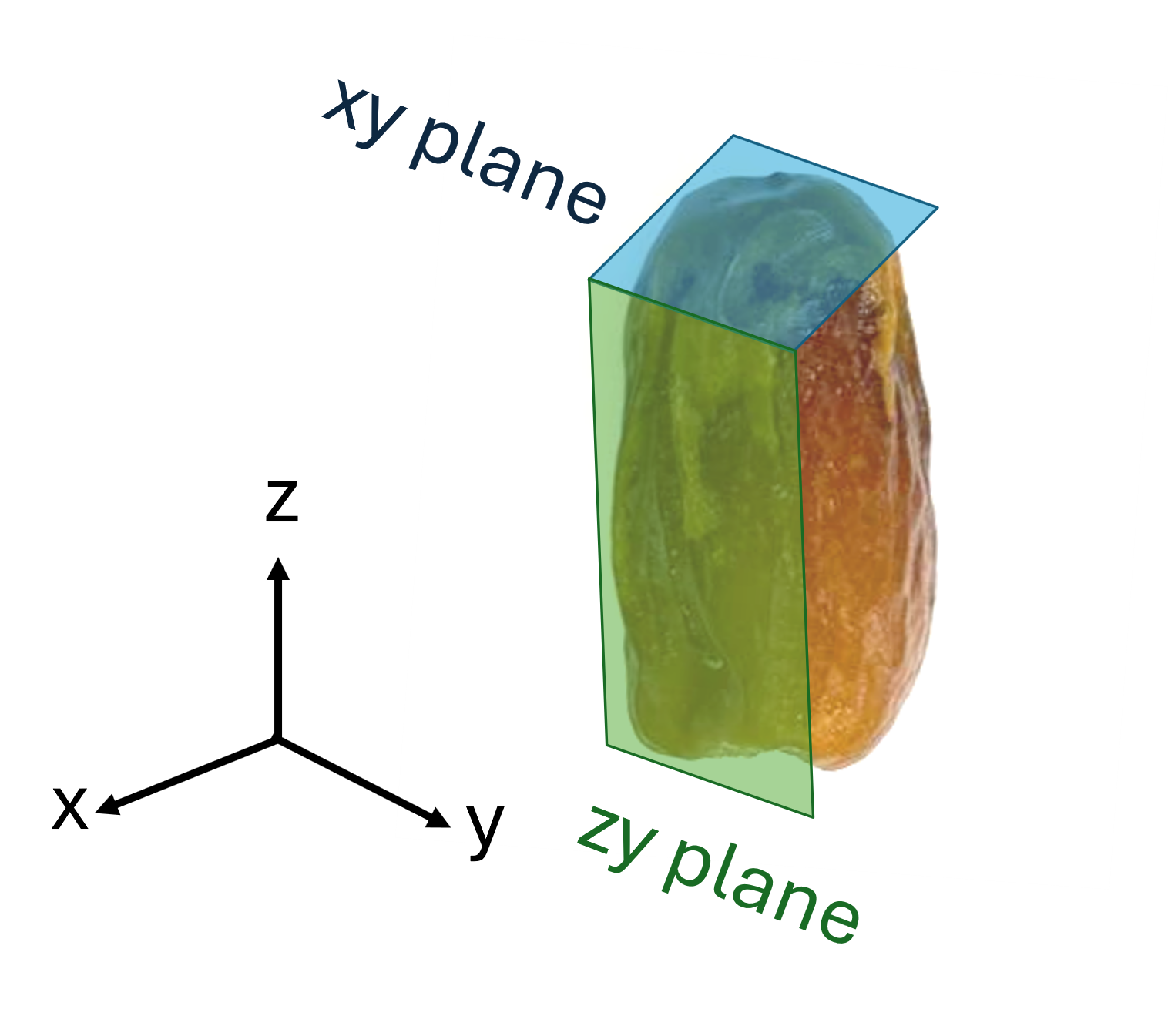}

    \caption{ Coordinate definition for date-fruit dimensions. 
}
    \label{fig:datexyz}
    \end{figure}

To provide standardized inputs for the V2F framework, we define a right-handed $x-y-z$ coordinate system centered on the fruit body, as illustrated in Fig. \ref{fig:datexyz}. The $z$-axis is aligned with the longitudinal stem-to-tip axis, representing the fruit height $Z_0$~(cm), while the $y$-axis corresponds to the maximum transverse diameter $Y_0$~(cm). For each sample, the cross-section of $A_0$~(cm$^2$) is measured on the $zy$-plane. Assuming a symmetric geometry about the $z$-axis ($Y_0 \approx X_0$), we estimate the fruit volume $V_0$~(cm$^3$) using the ellipsoidal approximation
\begin{equation}
V_0 = \frac{2}{3} A_0 \cdot Y_0.
\end{equation} 

Each sample corresponds to a single fruit compression/grasp instance. Let the target be the measured yield-level engineering stress $\sigma$ (kPa), denoted in our dataset as \texttt{stress\_y\_kPa}. We define the input vector $x \in \mathbb{R}^{d}$ by concatenating physical descriptors and categorical condition indicators
\begin{equation}
x =
\Big[
Y_0,\ Z_0,\ A_0,\ V_0,\ \mathrm{onehot}(\texttt{DT}),\ \mathrm{onehot}(\texttt{HS}),\ \epsilon_y
\Big],
\label{eq:input_vector}
\end{equation}
 where $\mathrm{onehot}(\texttt{DT}) \in \{0,1\}^{2}$ encodes the cultivar identity across three varieties with one category dropped as
\begin{equation}
\texttt{DT} \in \{\texttt{Ajwa},\; \texttt{Barhi},\; 
                  \texttt{Sagai}\},
\end{equation}
yielding two binary indicator columns $[\mathbf{1}_{\texttt{Barhi}},\; \mathbf{1}_{\texttt{Sagai}}]$, with \texttt{Ajwa} as the reference category, and $\mathrm{onehot}(\texttt{HS}) \in \{0,1\}^{2}$ encodes the hydration level across three conditions as
\begin{equation}
\texttt{HS} \in \{\texttt{Dried},\; \texttt{Moistened},\; 
                  \texttt{Normal}\},
\end{equation}
yielding two binary indicator columns $[\mathbf{1}_{\texttt{Moistened}},\; \mathbf{1}_{\texttt{Normal}}]$, with \texttt{Dried} as the reference category. Finally, $\varepsilon_y >0$ is 
 the cultivar-specific elastic strain limit obtained from
 experimental data. This strain corresponds to the maximum recoverable deformation prior to the onset of permanent tissue damage and serves as a biomechanical prior for force prediction.
\subsection{Vision Model Architecture}
\label{sec:ex3}
 
We employ an AI-based computer vision model for object detection with advanced image processing algorithms to track and quantify dimensional during experiment. The model and image processing are implemented in \texttt{Python} (v3.10.11) ~\cite{python310} with the \texttt{PyTorch} ~\cite{paszke2019pytorch} and \texttt{OpenCV}~\cite{opencv_library} packages to develop an autonomous system to be applicable for future systems.  
 
We select the Mask Region-Based Convolutional Neural Network model with Residual Network with 50 layers Feature Pyramid Network (MaskR-CNN with ResNet-50 FPN) backbone for date fruit detection and segmentation~\cite{hassan2022mask}. We fine-tune and train the model on our dataset to identify and segment date fruits in both static images and video sequences. Also, we utilize Common Objects in Context (COCO) pre-trained weights to accelerate the training for binary classification to label objects as background or date fruit. 
 
Additionally, we configure the region of interest heads with a Non-Maximum Suppression (NMS) threshold and a detection score threshold to optimize detection sensitivity while maintaining precision.  We incorporate monitoring protocols into the training, including early stopping based on validation loss, periodic checkpointing, and visualization of predictions. We evaluate training progress through detailed loss breakdown analysis covering classification, bounding box regression, mask prediction, objectness, and region proposal network box regression components~(Table~\ref{tab:maskrcnn_config}).
        
        \begin{table}[htbp]
        \centering
        \caption{MaskR-CNN Model Hyperparameter}
        \footnotesize
        \begin{tabular}{@{}p{4cm}p{4cm}@{}}
        \hline
        \textbf{Parameter Category} & \textbf{Configuration Details} \\
        \hline
        Training epochs & 100 \\
        Optimizer & Stochastic Gradient Descent \\
        Initial learning rate & 0.001 \\
        Momentum & 0.9 \\
        Weight decay & 0.0005  \\
        Batch size & 2 \\
        Mask loss weighting & 1.5  \\
        Early stopping patience & 5 epochs \\
        NMS threshold & 0.4 \\
        Detection score threshold & 0.3 \\
        \hline
        \end{tabular}
        \label{tab:maskrcnn_config}
        \end{table}

        To improve measurement accuracy, we implement an \textit{Active Contour Algorithm} that refines the initial model segmentation contour. The purpose of this algorithm is to ensure that the contour surrounds only the detected date fruit boundaries within each frame. The algorithm employs grayscale conversion with contrast enhancement using Contrast-Limited Adaptive Histogram Equalization (CLAHE)~\cite{zuiderveld1994clahe}. CLAHE is a local method that enhances image contrast by redistributing pixel intensity values within small regions while preventing the amplification of noise through contrast limiting. 
        The algorithm then incorporates the contours detected in previous frames to ensure temporal consistency throughout the video sequence, providing a logical correlation between detected object motions. Finally, the output refinement contour is constrained to the initial model contour's boundaries. 

For dimensional measurements, we use a ZED stereo camera~\cite{zed_camera} to capture images and videos. The dimensions are then extracted from the model’s detected contour.

\subsection{Vision Model Performance}

\begin{figure}[!htbp]
            \centering
            \subsection*{\small\normalfont (a) Epoch 1}
            \begin{minipage}[t]{0.49\linewidth}
                \centering
                \includegraphics[width=\linewidth]{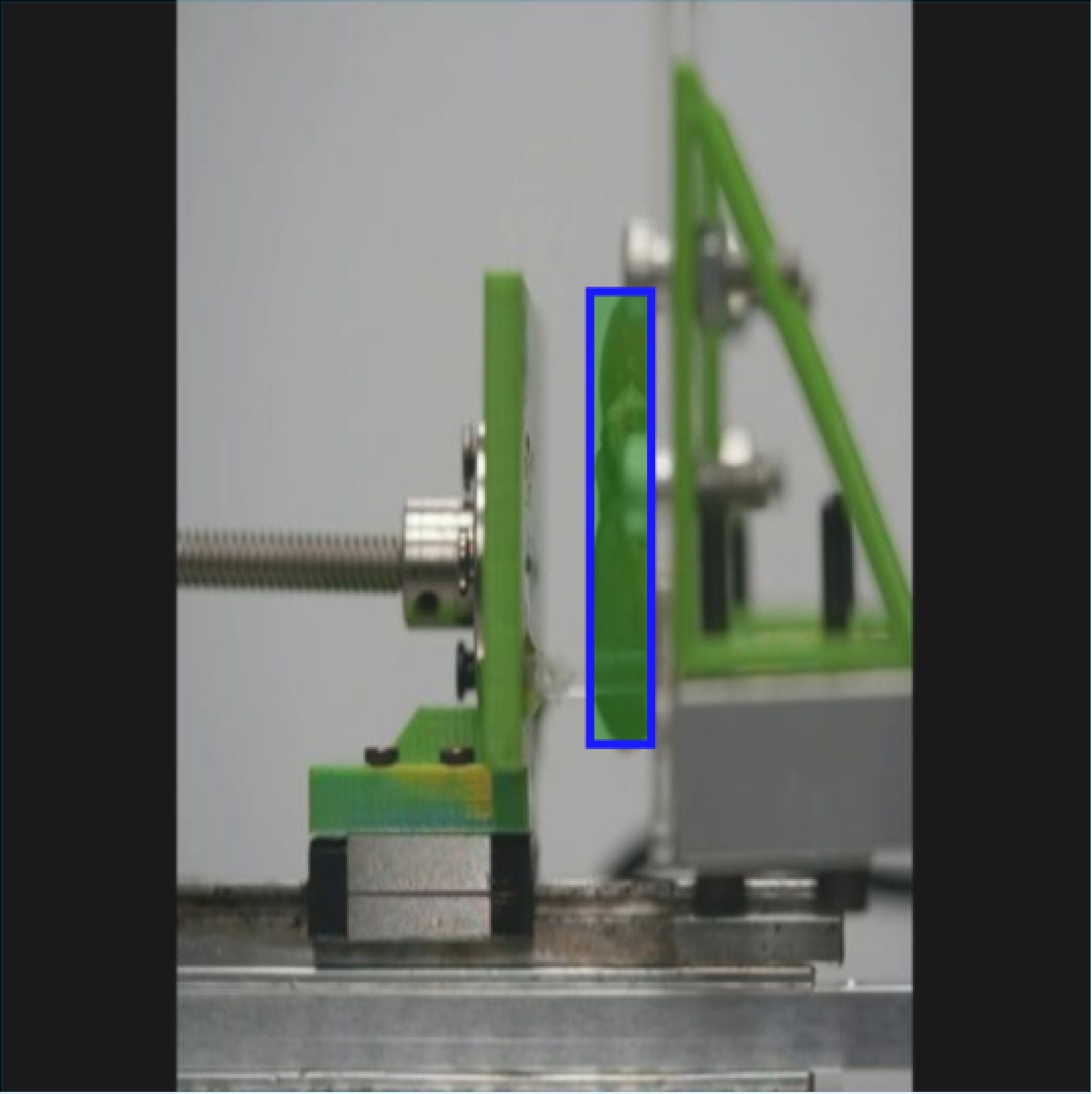}     
                {\small Ground Truth}
            \end{minipage}
            \hfill
            \begin{minipage}[t]{0.49\linewidth}
                \centering
                \includegraphics[width=\linewidth]{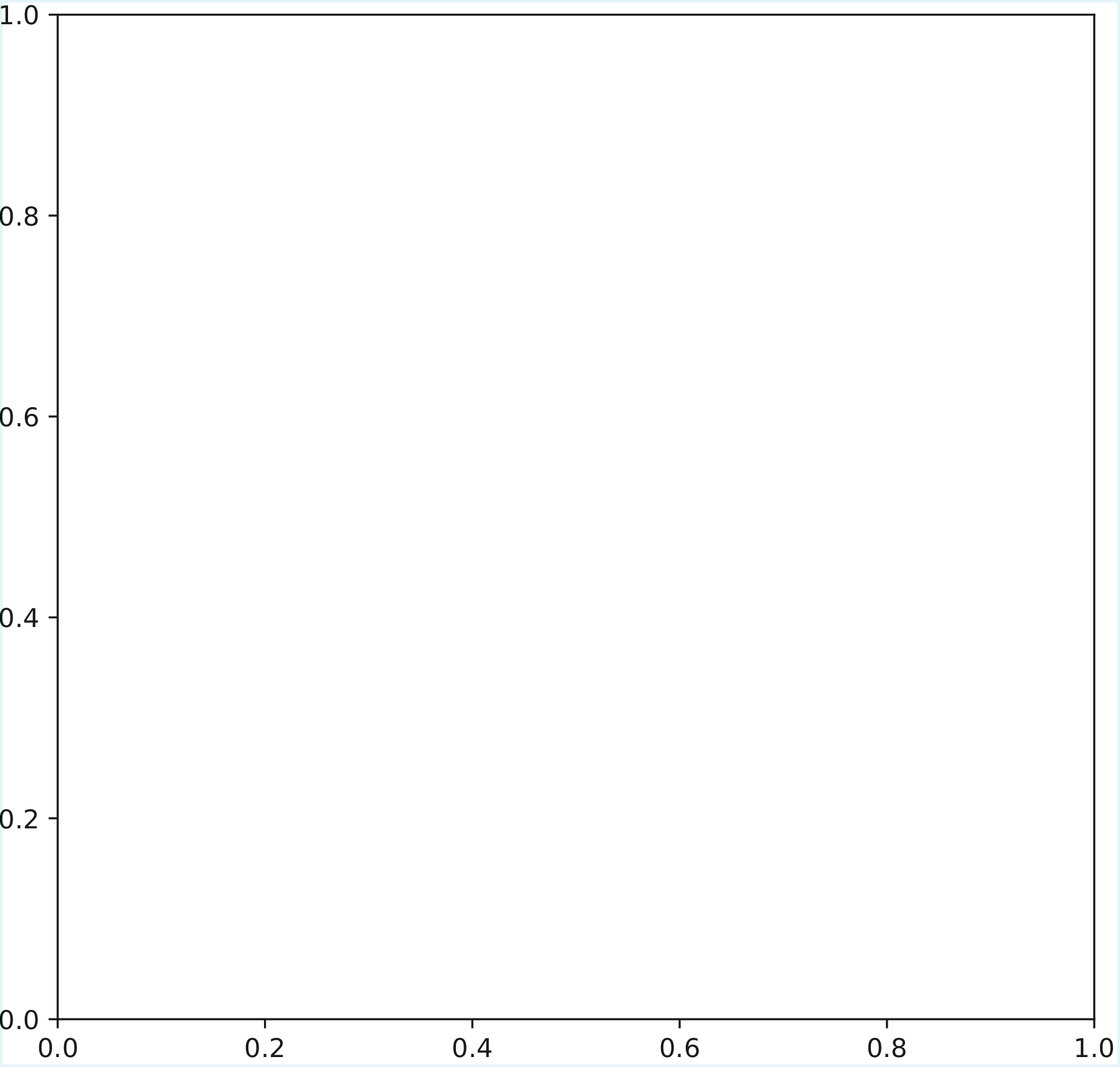}
                {\small Prediction}
            \end{minipage}
            \subsection*{\small\normalfont (b) Epoch 5}
            \begin{minipage}[t]{0.49\linewidth}
                \centering
                \includegraphics[width=\linewidth]{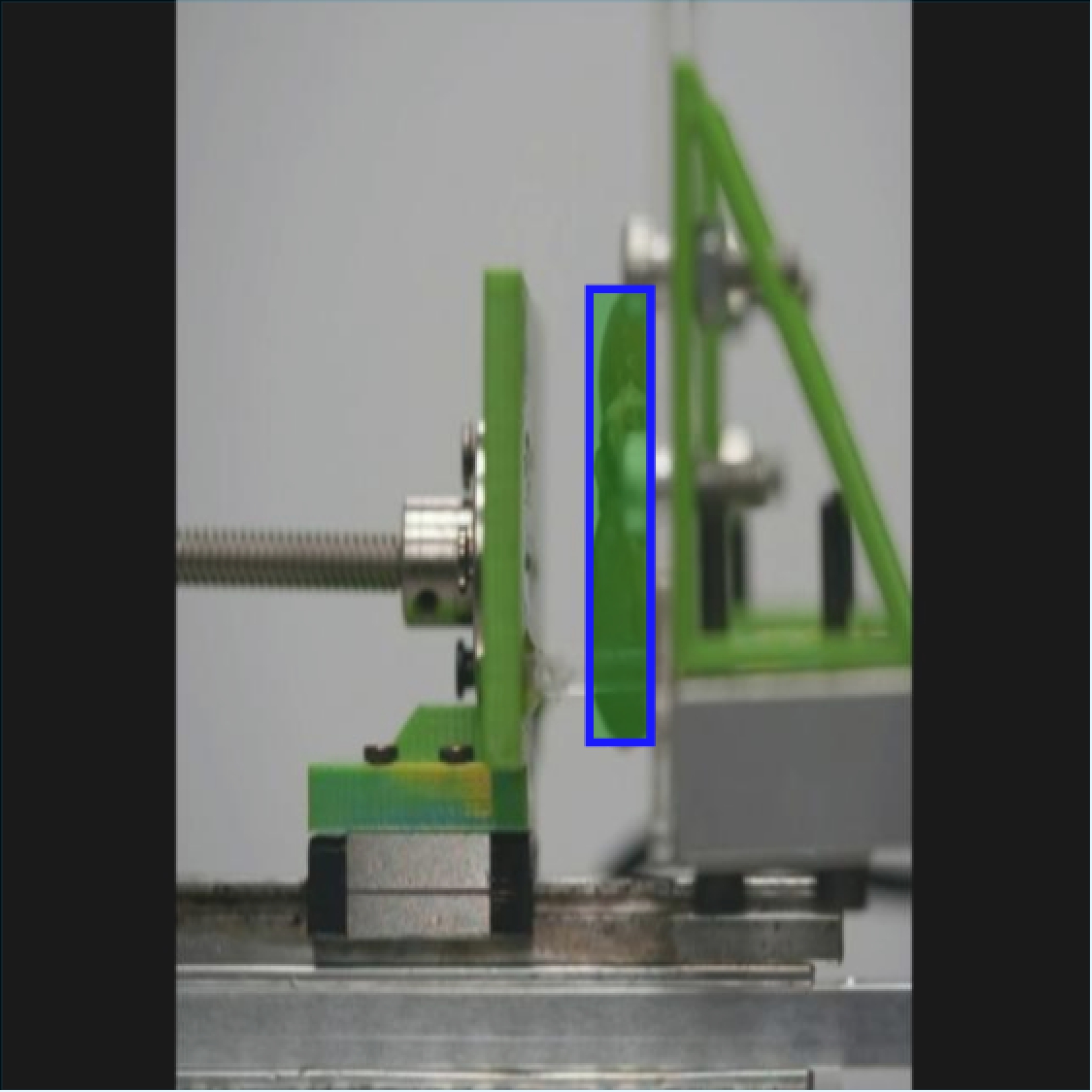}
                {\small Ground Truth}
            \end{minipage}
            \hfill
            \begin{minipage}[t]{0.49\linewidth}
                \centering
                \includegraphics[width= 6.2cm ,height= 4.4cm, trim=500 9.5 10 25, clip]{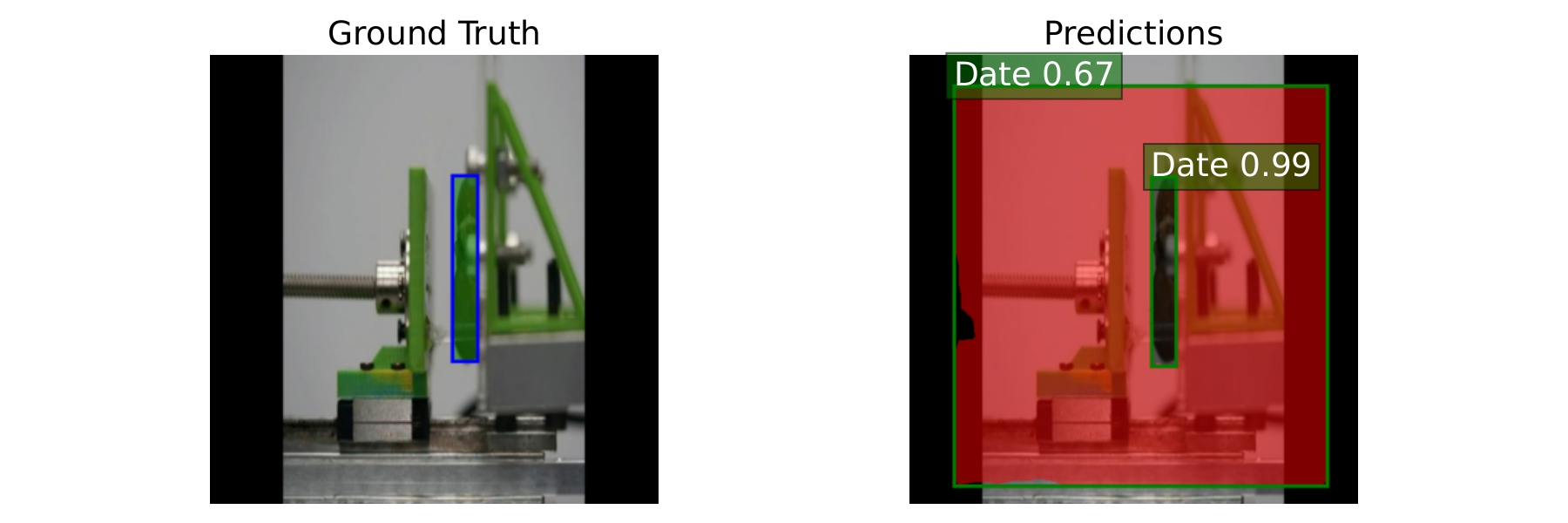}
                {\small Prediction}
            \end{minipage}
            \subsection*{\small\normalfont (c) Epoch 10}
            \begin{minipage}[t]{0.49\linewidth}
                \centering
                \includegraphics[width=\linewidth]{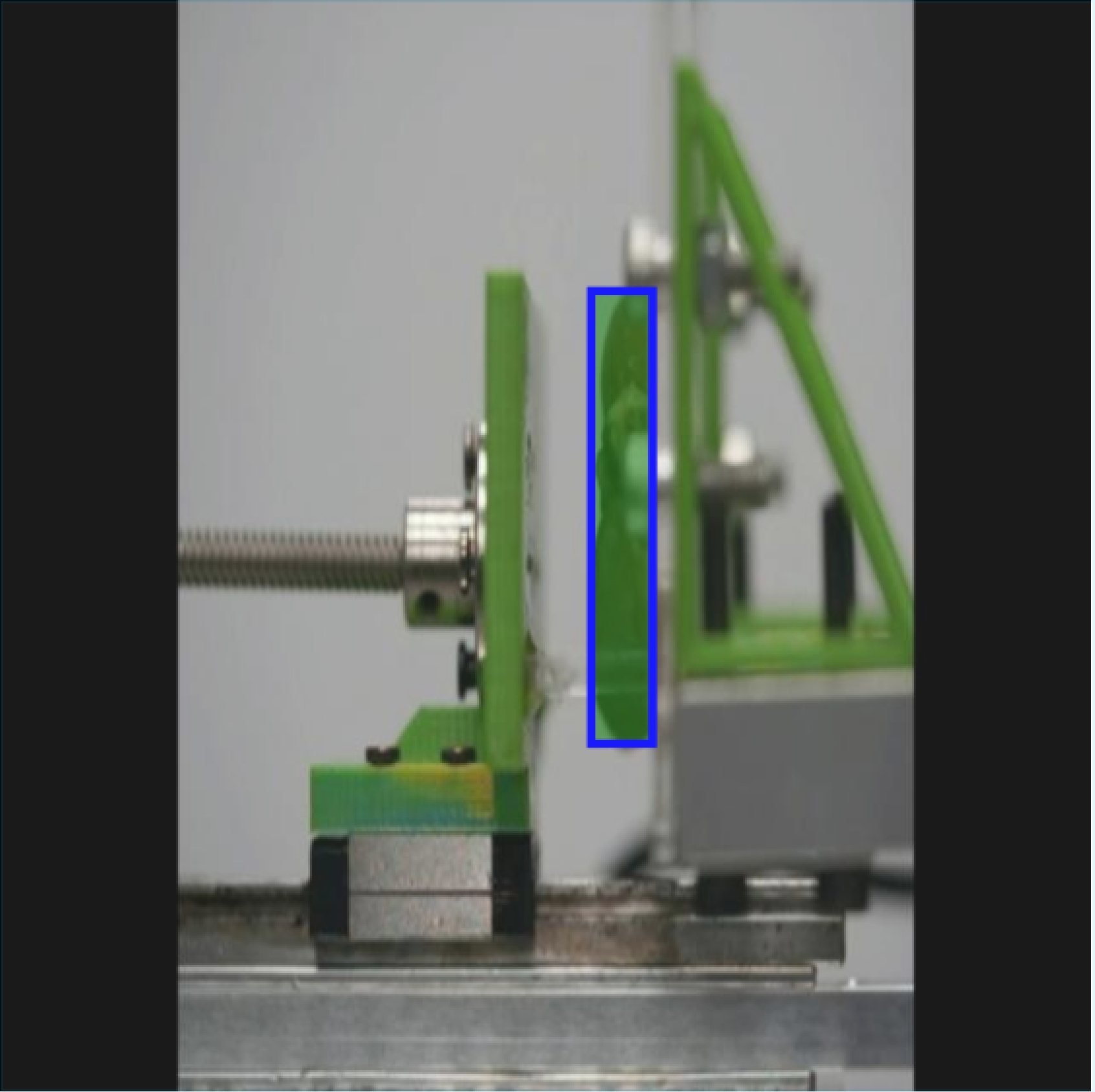}
                {\small Ground Truth}
            \end{minipage}
            \hfill
            \begin{minipage}[t]{0.49\linewidth}
                \centering
                \includegraphics[width= 6.2cm ,height= 4.4cm, trim=500 9.5 10 25, clip]{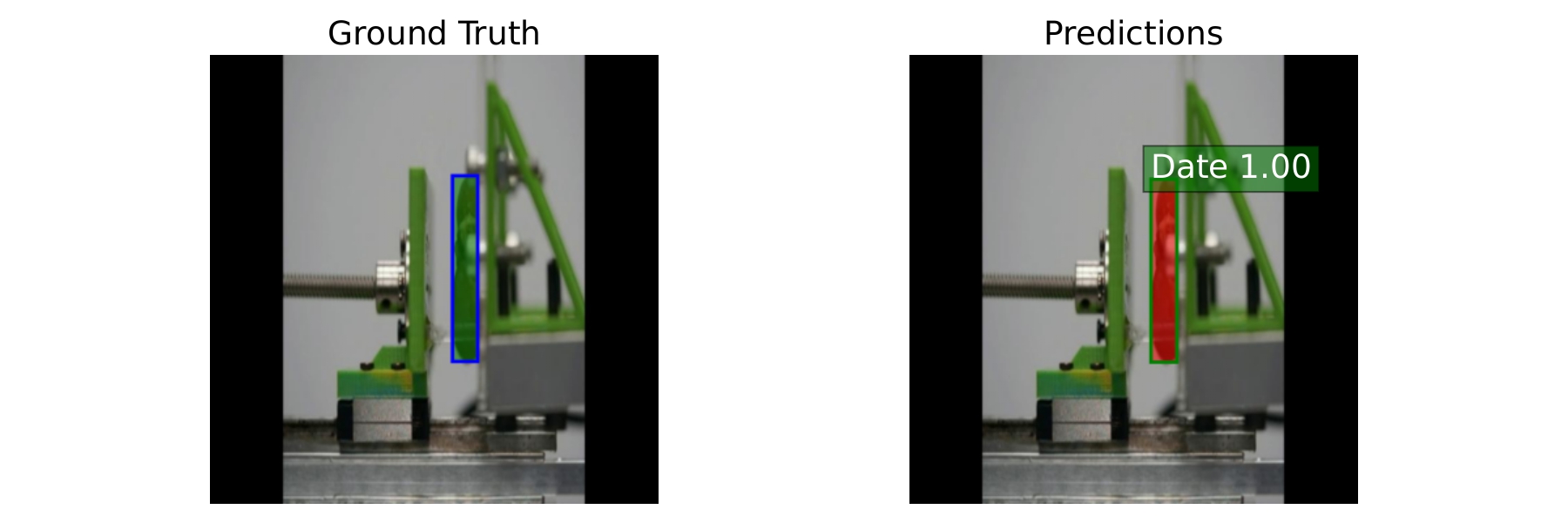}
                {\small Prediction}
            \end{minipage}
            \caption{
                Model predictions at different training epochs,
                The red boxes indicate detected objects with the green box displays the corresponding accuracy score. 
            }
            \label{fig:predictions-grid}
        \end{figure}

\begin{figure}[!htbp]
          \centering
          
            \includegraphics[
            height= 3.5cm
            , trim=120 80 120 90, clip]{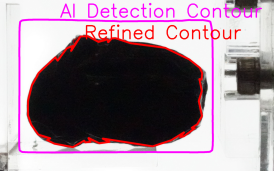}
            \caption{A frame illustrating the date fruit, lighting setup. The Mask R-CNN detection contour is shown in pink, while the refined active-contour boundary is shown in red.}
        
          \label{fig:area-and-raw}
        \end{figure}

The primary performance metric is classification accuracy, defined as the ratio of correctly classified instances to the total number of predictions. In epoch 1 shows early segmentation with no prediction above 0.5 accuracy, providing empty frame. The developed Mask R-CNN model achieved an initial accuracy of 0.67 at epoch 5, primarily due to background misclassifications. As training progressed, false positives and segmentation errors were progressively reduced. By epoch 10, the model reached stable predictions with an accuracy of 1.0, enabling precise measurement of fruit geometry.
As illustrated in Fig.~\ref{fig:predictions-grid}, the model demonstrates robust detection performance, including accurate segmentation of compressed dates. This reliable segmentation directly supports accurate residual deformation calculations. Overall, the results indicate that automated segmentation can effectively measure date fruit dimensions, overcoming the limitations of manual caliper-based methods reported in prior studies \cite{FCHSA,Mahawar2017}.
        With reliable segmentation provided by the Mask R-CNN model, the next step in the pipeline is \textit{Active Contour Algorithm}. This algorithm extracts a refinement contour and outputs the physical properties measured in pixels. Existing studies focus mainly on classification or simple feature extraction of single date fruits~\cite{altaheri2019date,su14106339}, whereas our work advances this area by providing active-contour segmentation, precise geometric measurement, and improved handling of the irregular textures and natural asymmetry characteristic of date fruits~(Fig.~\ref{fig:area-and-raw}). 

\begin{table}[t]
    \centering
    \caption{Baseline model comparison under GroupKFold cross-validation(CV).
    Train and validation RMSE are reported in kPa. CV $R^2$ mean and 
    standard deviation are computed across five folds. Models with 
    negative CV of $R^2$ perform worse than a constant mean predictor.}
    \label{tab:baseline_comparison}
    \renewcommand{\arraystretch}{1.15}
    \begin{tabular}{lcccccc}
        \hline
        \textbf{Model} & 
        \textbf{\shortstack{Val\\RMSE}} & 
        \textbf{\shortstack{Val\\$R^2$}} & 
        \textbf{\shortstack{CV $R^2$\\Mean}} & 
        \textbf{\shortstack{CV $R^2$\\Std}} \\
        \hline
        Baseline (Mean)      & 1.548 & $-$0.095 & $-$0.032 & 0.011 \\
        Baseline (Median)    & 1.480 & $-$0.001 & $-$0.064 & 0.029 \\
        Linear Regression   & 1.488 & $-$0.012 & 0.054 & 0.077 \\
        Ridge               & 1.469 & 0.014 & 0.126 & 0.029 \\
        Lasso                 & 1.548 & $-$0.095 & $-$0.032 & 0.011 \\
        ElasticNet          & 1.466 & 0.017 & 0.076 & 0.048 \\
        Random Forest       & 1.498 & $-$0.025 & 0.173 & 0.056 \\
        Gradient Boosting    & 2.063 & $-$0.945 & $-$0.080 & 0.129 \\
        Extra Trees          & 1.398 & 0.106 & 0.166 & 0.058 \\
        SVR (RBF)             & 1.494 & $-$0.019 & 0.069 & 0.044 \\
        SVR (Linear)        & 1.428 & 0.068 & 0.126 & 0.070 \\
        KNN                 & 1.561 & $-$0.113 & $-$0.012 & 0.122 \\
       \textbf{The proposed model} & \textbf{1.1345} & \textbf{0.7407} & \textbf{0.6986} & \textbf{0.0828} \\
        \hline
        \multicolumn{5}{l}{\footnotesize Bold entries indicate the best 
        validation $R^2$ per column.}\\
    \end{tabular}
\end{table}
\begin{figure}[!htbp]


        \includegraphics[width=\linewidth]{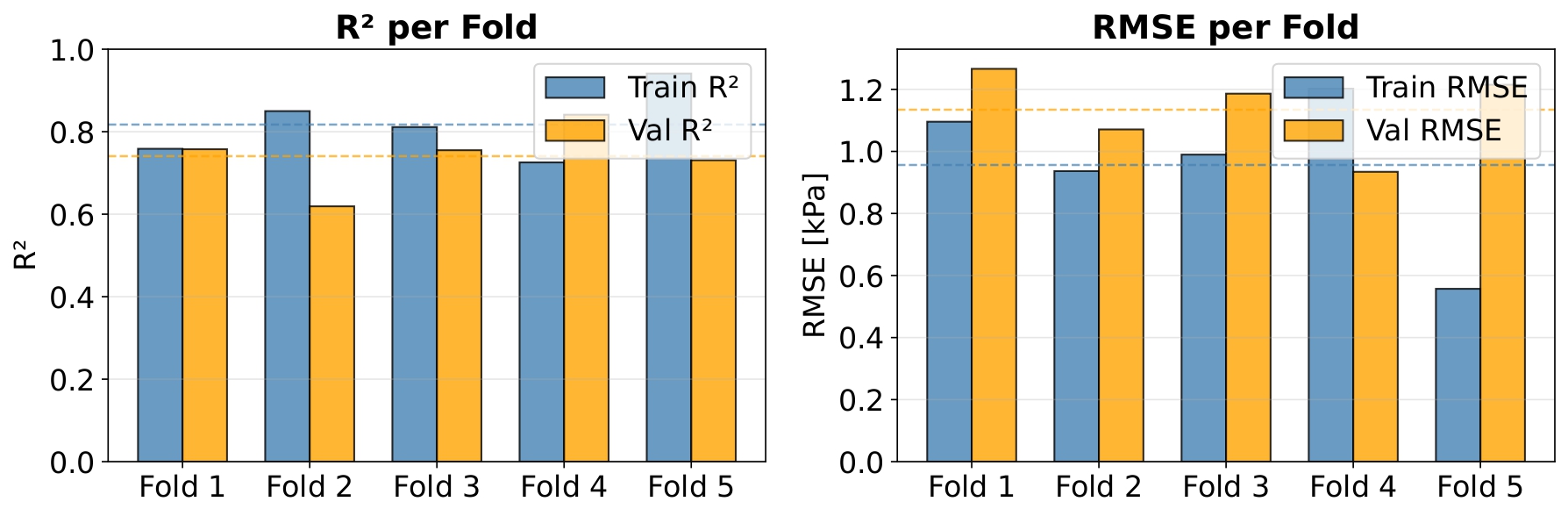}
    \caption{Per-fold cross-validation performance under GroupKFold partitioning, $R^2$ scores and RMSE (kPa), dashed lines indicate the mean across folds (Train: blue, Val: orange).}
    \label{fig:fold_summary}
\end{figure}

\subsection{Stress Prediction ML Model  Theoretical Formulation}
To model the complex biomechanical response of fruits, we propose a hybrid framework that augments a classical analytical baseline with a learnable residual term. We first compute an analytical baseline stress, $\sigma_{\mathrm{Hertz}}$, derived from Hertzian contact theory,
We estimate mechanical parameters $(\hat{E},\hat{\nu})$ from physical features via regression to define the effective modulus $\hat{E}^{*} = \hat{E}/(1-\hat{\nu}^{2})$. The equivalent fruit radius is approximated as $R_{\mathrm{eq}} = Y_0$, and indentation is modeled as $\delta = ( \epsilon_y/100 )\,Y_0$. The resulting baseline is
\begin{equation}
\hat\sigma_{\mathrm{Hertz}}(x) = \frac{4}{3}\,\hat{E}^{*}\,\sqrt{R_{\mathrm{eq}}}\,\frac{\delta^{3/2}}{A_0}.
\label{eq:hertz_stress}
\end{equation}
Since biological fruits deviate from ideal elastic assumptions due to viscoelasticity and structural heterogeneity, we decompose the predicted stress into this Hertz term plus a learnable residual $r_{\phi}(x)$
\begin{equation}
\hat\sigma = \hat\sigma_{\mathrm{Hertz}}(x) + r_{\phi}(x) + \eta,
\label{eq:decomposition}
\end{equation}
where $\hat\sigma_{\mathrm{Hertz}}(x)$ is the predicted Hertz's stress, $r_{\phi}(x)$ 
is the learned residual with network parameters $\phi$, and $\eta$ 
captures measurement noise. To enforce consistency between the data-driven predictions and the underlying contact mechanics, we formulate a physics-guided loss function, $\mathcal{L}$, that combines a supervised data loss with the residual of a learnable effective Hertzian contact model. The overall training objective is defined as
\begin{equation}
\mathcal{L} =
(1-\alpha)\,\mathcal{L}_{\mathrm{data}}\!\left(\hat{r}_{\phi},\,r\right)
+
\alpha\,\mathcal{L}_{\mathrm{phys}}\!\left(\hat{\sigma}(x),\,\hat{\sigma}_{\mathrm{Hertz}}(x)\right),
\label{eq:loss_total}
\end{equation}
where $\mathcal{L}_{\text{data}}$ penalizes the mismatch between the predicted residual stress $\hat{r}_\phi$ and the ground-truth residual $r$, $\mathcal{L}_{\text{phys}}$ penalizes deviations of the total predicted stress $\hat{\sigma}$ from the Hertz's contact stress $\hat{\sigma}_{\mathrm{Hertz}}$, and $\alpha \in [0,1]$ controls the trade-off between fitting the residual to data and maintaining consistency with the Hertzian contact model. Setting $\alpha = 0$ reduces to a pure data-driven residual, 
while $\alpha = 1$ enforces full physics consistency. This physics-guided residual learning encourages the network to produce stress fields that remain close to the analytical Hertz solution while still capturing unmodelled effects, leading to highly accurate predictions of the final grasping force $F_{grasp}$.


\subsection{Validation and Generalization of the Proposed Method}
To assess the robustness of the V2F framework across diverse conditions, we define a group label $g =~(\text{date\_type, hydration\_state})$ for each sample. We implemented a five-fold GroupKFold cross-validation (CV) strategy, ensuring that samples from the same cultivar–state group are never split across training and validation sets. This rigorous partitioning, illustrated in Fig. \ref{fig:fold_summary}, forces the model to generalize its predictions to entirely unseen date varieties.
As shown in Table \ref{tab:baseline_comparison}, the proposed model significantly outperforms standard baselines, achieving a mean validation $R^2$ of 0.74 and an RMSE of 1.13 kPa. While Fold 5 (Sagai Dried/Normal) presented a greater generalization challenge ($R^2 = 0.69$), the consistently low training RMSE (0.56 kPa) suggests that while internal features are learned well, high inter-group variability requires our hybrid approach to maintain stability. Overall, the consistent performance confirms that the model provides physically reasonable force set-points even when encountering novel cultivar-condition combinations.


\section{Experimental Validation}
\label{sec:Exp_validation}

In this section, we evaluate the performance of the proposed Vision-to-Force (V2F) framework through physical grasping experiments. The primary objective is to verify if the AI-predicted force $F_{\text{pred}}$ is sufficient for stable manipulation while remaining below the bioyield threshold to prevent permanent damage.


\subsection{Experimental Procedure}
The validation protocol follows a three-stage sequence, see Fig.~\ref{fig:Exp_pipeline}:
\begin{enumerate}
    \item The ZED camera captures the scene. The vision model performs fruit segmentation to calculate the projected area $A_p$. This value, alongside the variety and ripeness state, is fed into the V2F model.
    \item The model generates a target force $F_{\text{pred}}$. The Robotiq 2F-85 gripper \cite{robotiq_adaptive_grippers} is then actuated until the integrated load cell registers the target set-point.
    \item A standard pick-and-place task is executed. Upon release, the vision system re-scans the fruit to quantify the residual deformation,
\begin{equation}
\delta_{\mathrm{res}} = L_{\mathrm{after}} - L_0,
\end{equation}
where $L_0$ is the initial fruit dimension before grasping and $L_{\mathrm{after}}$ is the corresponding dimension measured after the fruit is released and allowed to recover. The residual deformation serves as a quantitative indicator of permanent mechanical damage.
\end{enumerate}

\subsection{Experimental Results and Discussion}
\begin{figure}[!htbp]
    \centering
      \includegraphics[width=\linewidth]{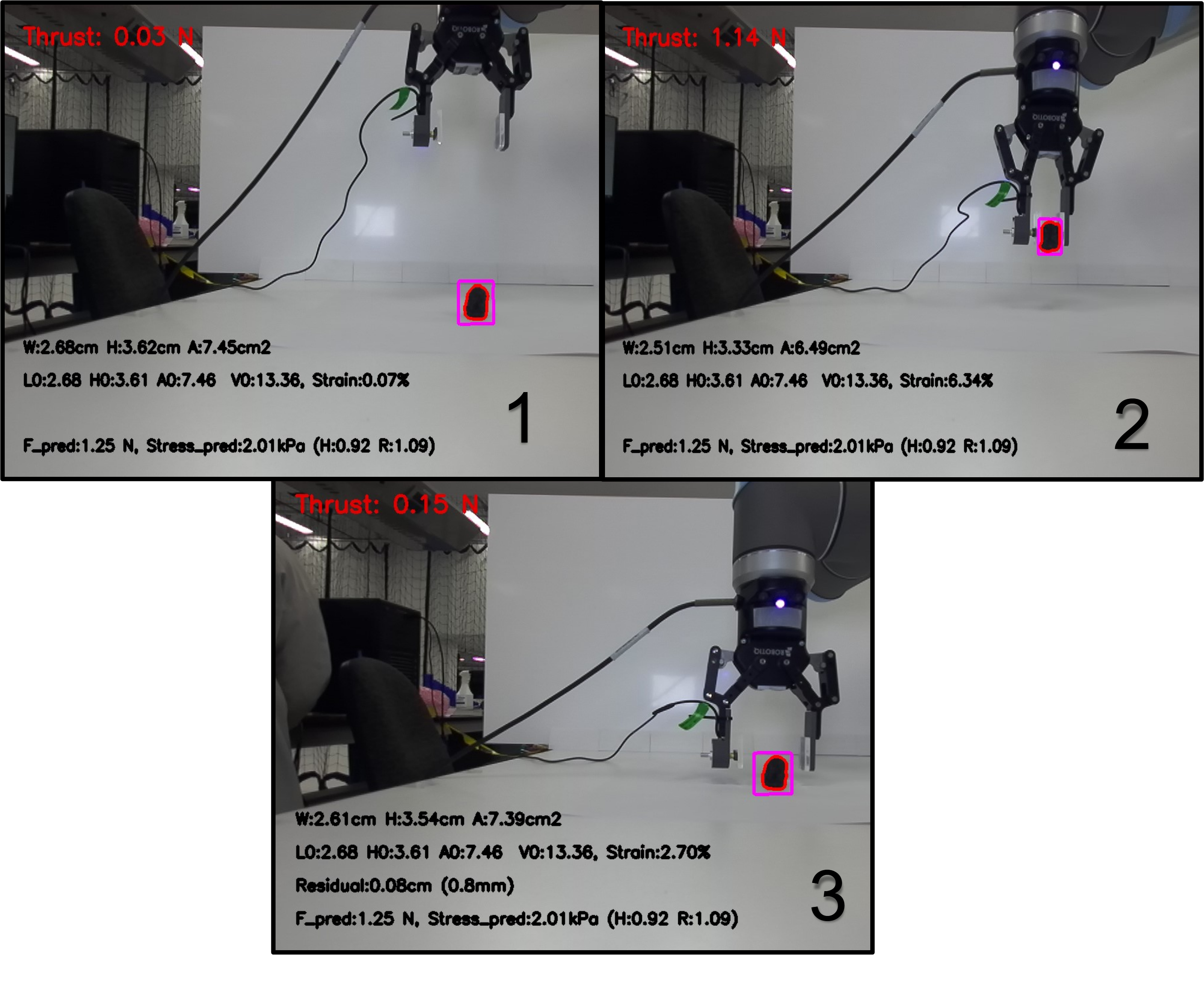}
    \caption{Experiment workflow, (1) identify the fruit area and needed force, (2) grip and move the date with required force, (3) release the date and calculate the residual deformation.}
    \label{fig:Exp_pipeline}
\end{figure}

The framework was evaluated on multiple date cultivars, specifically \textit{Ajwa}, \textit{Barhi}, and \textit{Sagai}. 
All experiments were conducted at the natural (normal) hydration state, representing the most common commercial harvesting condition.
The experimental workflow is illustrated in Fig.~\ref{fig:Exp_pipeline}, while the quantitative results are summarized in Table~\ref{tab:results}.

\begin{table}[h]
    \centering
    \caption{Experimental Validation Results for V2F Grasping}
    \label{tab:results}
    \begin{tabular}{lccc}
    \hline
    \textbf{Date Variety} & \textbf{Area} & \textbf{Predicted Force } & \textbf{Residual} \\
     \textbf{Unit} & \textbf{$ mm^2$} &\textbf{$N$} & \textbf{$\Delta mm$} \\ \hline
    Ajwa & 7.473 & 1.25 & 0.8 \\
    Barhi & 7.123 & 0.60 & 0.6 \\ 
    Sagei & 7.126 & 0.30 & 0.4  \\ \hline
    \end{tabular}
\end{table}

The predicted grasp force varied according to fruit morphology and cultivar, ranging from $0.30$~N for \textit{Sagai} to $1.25$~N for \textit{Ajwa}. Despite this variation, all grasping trials were completed successfully without fruit dropping or visible bruising. The measured post-release residual deformation remained below $1$~mm for all cultivars, with values of $0.8$~mm, $0.6$~mm, and $0.4$~mm for \textit{Ajwa}, \textit{Barhi}, and \textit{Sagai}, respectively. These results indicate that the predicted forces maintained the fruit within the elastic deformation regime established during the mechanical characterization, thereby preventing permanent tissue damage.

To verify that the predicted grasp forces ensure both damage-free and stable manipulation, grasp stability was further evaluated using the Coulomb friction model. For a two-finger gripper, slip-free transport requires
\begin{equation}
2\mu_s\hat{F}_n \ge m_f(g+a_{\max}),
\label{eq}
\end{equation}
where $\mu_s$ is the coefficient of static friction, $\hat{F}_n$ is the predicted normal grasp force, $m_f$ is the fruit mass, $g$ is the gravitational acceleration, and $a_{\max}$ is the maximum transport acceleration. Using a conservative coefficient of static friction of $\mu_s=0.3$ and an average fruit mass of $9$~g~\cite{Ghonimy2025PhysicoMechanical}, the minimum normal force required for static slip-free grasping is approximately $0.15$~N. This value is well below the minimum force predicted by the proposed framework ($0.30$~N for \textit{Sagai}), indicating that all predicted grasp forces satisfy the theoretical slip-free condition with a safety margin of at least two. Consistent with this analysis, no observable slip or fruit loss occurred during any pick-and-place experiment.

The experimental results demonstrate that the proposed V2F framework simultaneously achieves damage-free handling and reliable grasp stability. Unlike conventional grasping approaches based on fixed force heuristics or reactive tactile sensing, the proposed method predicts fruit-specific grasp forces directly from pre-contact visual observations while incorporating Hertzian contact mechanics through a physics-guided residual learning framework. Consequently, the robot adapts its grasp before contact, eliminating exploratory squeezing while maintaining stable manipulation across different fruit cultivars.

These results substantiate the dual-objective design of the framework: 
(i) damage-free handling and (ii) reliable grasp stability, validating 
the effectiveness of embedding physics-guided residual learning within 
a vision-based force prediction pipeline.

\section{Conclusion}
\label{sec:conclusion}

This work presents a vision-informed Vision-to-Force (V2F) framework that predicts safe grasping forces for robotic handling of date fruits directly from visual features and minimal contextual metadata. The model is grounded in a mechanical characterization study and integrates a Hertz analytical baseline within a residual neural network to ensure physically reasonable stress predictions across different cultivar and hydration states. The framework combines Mask R-CNN segmentation, active-contour refinement, and physics-informed residual learning to estimate fruit geometry and safe stress levels. Using GroupKFold cross-validation with cultivar-state groups held out, the model achieves a mean validation 
$R^2 \approx 0.7$. Robotic experiments on a manipulator with a force-sensor-equipped gripper demonstrate that the predicted forces enable stable manipulation while maintaining residual deformation below 1 mm, indicating operation within the elastic regime without bruising.
\noindent
More broadly, this approach demonstrates how non-contact vision can replace slow exploratory tactile sensing, which risks micro-bruising and reduces handling throughput. Future work will explore closed-loop integration with tactile feedback to handle rare out-of-distribution cases where vision-only predictions may be insufficient.

\bibliographystyle{IEEEtran}
\bibliography{refer2.bib}

\end{document}